\documentclass[10pt,twocolumn,letterpaper]{article}

\usepackage{iccv}              

\definecolor{iccvblue}{rgb}{0.21,0.49,0.74}

\usepackage[pagebackref,breaklinks,colorlinks,allcolors=iccvblue]{hyperref}

 \usepackage{multirow}
\usepackage{amssymb}
\usepackage{times}
\usepackage{soul}
\usepackage{url}
\usepackage{wrapfig}
\usepackage[utf8]{inputenc}
\usepackage{graphicx}
\usepackage{amsmath}
\usepackage{amsthm}
\usepackage{booktabs}
\usepackage{algorithm}
\usepackage{algorithmic}
\usepackage[switch]{lineno}
\usepackage{pifont}
\usepackage[accsupp]{axessibility}  

\def\paperID{9710} 
\def\confName{ICCV}
\def\confYear{2025}

 \title{Rethinking Multi-Modal Object Detection from the Perspective of Mono-Modality Feature Learning}
\author{Tianyi Zhao$^{1}$\textsuperscript{*}, Boyang Liu$^{1}$\textsuperscript{*}, Yanglei Gao$^1$, Yiming Sun$^2$, Maoxun Yuan$^{1}$\textsuperscript{$\dagger$}, Xingxing Wei$^{1}$\textsuperscript{$\dagger$} \\
$^1$Institute of Artificial Intelligence, State Key Laboratory of Virtual Reality Technology and Systems,\\ Beihang University, 
$^2$Southeast University\\
{\tt\small ty\_zhao@buaa.edu.cn, by.liu2004@gmail.com, gaoyl04@buaa.edu.cn} \\ {\tt\small  sunyiming@seu.edu.cn, \{yuanmaoxun,xxwei\}@buaa.edu.cn}
}

\begin{document}
\maketitle
\footnotetext{*: Equal contribution. $\dagger$: Corresponding Author.} 

\begin{abstract}
Multi-Modal Object Detection (MMOD), due to its stronger adaptability to various complex environments, has been widely applied in various applications. Extensive research is dedicated to the RGB-IR object detection, primarily focusing on how to integrate complementary features from RGB-IR modalities. 
However, they neglect the mono-modality insufficient learning problem, which arises from decreased feature extraction capability in multi-modal joint learning. This leads to a prevalent but unreasonable phenomenon\textemdash Fusion Degradation, which hinders the performance improvement of the MMOD model.
Motivated by this, in this paper, we introduce linear probing evaluation to the multi-modal detectors and rethink the multi-modal object detection task from the mono-modality learning perspective.
Therefore, we construct a novel framework called \textbf{M$^2$D-LIF}, which consists of the \textbf{M}ono-\textbf{M}odality \textbf{D}istillation (\textbf{M$^2$D}) method and the \textbf{L}ocal \textbf{I}llumination-aware \textbf{F}usion (\textbf{LIF}) module.
The M$^2$D-LIF framework facilitates the sufficient learning of mono-modality during multi-modal joint training and explores a lightweight yet effective feature fusion manner to achieve superior object detection performance. 
Extensive experiments conducted on three MMOD datasets demonstrate that our M$^2$D-LIF effectively mitigates the Fusion Degradation phenomenon and outperforms the previous SOTA detectors. The codes are available at \href{https://github.com/Zhao-Tian-yi/M2D-LIF}{https://github.com/Zhao-Tian-yi/M2D-LIF}.

\end{abstract}

\section{Introduction}
\label{sec:intro}
\begin{figure}[!t]
  \centering
    \includegraphics[width=\linewidth]{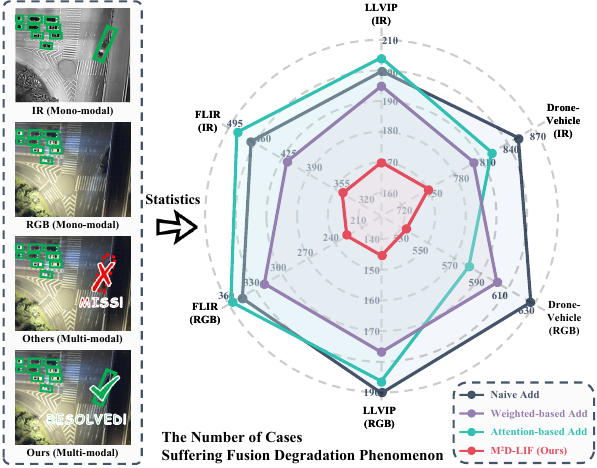} 
      \vspace{-0.5cm}
  \caption{Fusion Degradation phenomenon. (Left) An instance of this phenomenon. The object highlighted by the \textcolor{red}{red box} is detected by a mono-modal method but missed by the multi-modal method. (Right) The Fusion Degradation phenomenon statistics of three methods and our method across different datasets.}
         \vspace{-0.5cm}
    \label{fig:1}
\end{figure}
Recently, multi-modal object detection (MMOD) technology has been widely used in various safety-critical applications such as around-the-clock pedestrian detection for urban surveillance~\cite{jia2021llvip}, object detection for autonomous driving~\cite{person2019multimodal} etc. Since relying solely on visible images~\cite{zhang2022lane,xia2024vit} will render objects invisible under limited illumination, utilizing both visible (RGB) images and infrared (IR) images has been widely used as one of the effective solutions to achieve full-time object detection tasks. Specifically, RGB modality can provide the texture of objects in daylight, while infrared modality can provide the outline of objects under poor lighting conditions. By exploring the complementary information between the two modalities, RGB-IR object detection~\cite{li2024fd2,zhange2e} can aggregate the advantages of each modality to achieve robust visual perception. 

To design advanced fusion strategies for the MMOD task, previous works~\cite{liu1611multispectral,li2018multispectral,wolpert2020anchor} reveal that the feature-level fusion methods outperform both image-level and decision-level fusion methods. Building upon this finding, recent RGB-IR object detection methods mainly focus on designing complex feature fusion structures to address various challenges. For example, CSSA~\cite{cao2023multimodal} and ICAFusion~\cite{shen2024icafusion} introduce different attention mechanisms to better explore multi-modal complementarity. CALNet \cite{he2023multispectral} proposes the selected cross-modal fusion module to address the semantic conflict issue between different modalities. AR-CNN \cite{zhang2019weakly,zhang2021weakly} and TSFADet \cite{yuan2022translation,yuan2024improving} focus on solving the misalignment problem and attempt to improve object detection performance by aligning RGB and IR modality features. 
Although these methods have achieved encouraging object detection performance, they primarily focus on integrating complementary features from RGB-IR modalities while neglecting the mono-modality insufficient learning problem. This results in an unreasonable phenomenon that some objects can be detected by the mono-modal detector but missed by the corresponding multi-modal detector. We name this phenomenon as {Fusion Degradation} as illustrated in Figure~\ref{fig:1}. Besides, we statistics this phenomenon on three commonly used MMOD datasets and find that it is prevalent in different types of multi-modal detectors. Motivated by this, we rethink the multi-modal object detection task from the mono-modality learning perspective.

In this paper, we construct \textbf{M$^2$D-LIF}, an end-to-end framework for multi-modal object detection, to address the above problem. Different from the existing MMOD methods that design complex feature fusion modules, M$^2$D-LIF facilitates the sufficient learning of mono-modality during multi-modal joint training and explore a lightweight yet effective feature fusion manner to achieve superior object detection performance. Inspired by knowledge distillation \cite{hinton2015distilling}, we propose an  \textbf{M}ono-\textbf{M}odality \textbf{D}istillation (\textbf{M$^2$D}) method, which introduces the additional encoder pretrained on the mono-modality to distill the multi-modal encoders. To bridge the capability gap between the multi- and mono-modalities, we design the inner-modality and cross-modality distillation loss to jointly optimize the M$^2$D-LIF during training. In this way, we can ensure sufficient learning of the mono-modal encoder in the multi-modality joint training. Besides, according to our observations in Section~\ref{subsec: 3.1}, we propose a \textbf{L}ocal \textbf{I}llumination-aware \textbf{F}usion (\textbf{LIF}) module, which can dynamically set different weights for different illumination regions. The module ensures an explicitly complementary fusion to cooperate with M$^2$D method, thereby enhancing both the accuracy and efficacy of the multi-modal object detection. Figure~\ref{fig:1} shows that our framework can effectively solve the above problem.
Our contributions in this paper are highlighted as follows:

\begin{itemize}

    \item We introduce linear probing evaluation to the multi-modal detectors and identify the insufficient learning of mono-modality during training. To the best of our knowledge, it is the first time to rethink multi-modal object detection from a mono-modality learning perspective.
    
    \item We present M$^2$D-LIF, a pioneering method to improve mono-modality learning capabilities during multi-modal joint training, innovatively exploring a lightweight yet effective feature fusion framework for the MMOD task.
    
    \item Extensive experiments on the three MMOD datasets demonstrate that our M$^2$D-LIF outperforms the previous state-of-the-art detectors and can be used as an effective feature fusion method in the MMOD task.

\end{itemize}

\section{Related Works}

\subsection{RGB-Infrared Object Detection}

Due to the deep exploration of complementary features through CNNs, feature-level fusion has become a widely used approach for RGB-Infrared object detection. 
In early studies, the weighted-based fusion methods have been used as a simple and effective way to fuse different features. Li~\textit{et al.}~\cite{li2019illumination} proposed the first illumination-aware Faster R-CNN, which introduced illumination conditions into the RGB-Infrared object detection task. At the same time, Guan~\textit{et al.}~\cite{guan2019fusion} presented a multispectral pedestrian detection framework based on illumination-aware pedestrian detection and semantic segmentation. 
With the introduction of the attention mechanism, more and more attention-based fusion methods have been proposed to achieve complementary feature fusion. Fang~\textit{et al.}~\cite{qingyun2022cross} proposed the cross-modality attentive feature fusion (CMAFF) module, leveraging common-modality and differential-modality attentions. 
Concurrently, Cao~\textit{et al.}~\cite{cao2023multimodal} introduced the CSSA fusion module, which employs channel switching and spatial attention for feature fusion. Besides, CMX~\cite{zhang2023cmx} was proposed to cross-modal feature rectification and feature fusion with intertwining cross-attention. Furthermore, Shen~\textit{et al.}~\cite{shen2024icafusion} proposed a dual cross-attention transformers to model global feature interaction and capture complementary information between two modalities.

However, the above methods mainly focus on designing complex and parameter-intensive modules to achieve the so-called complementary fusion, while ignoring the insufficient representation of each modality caused by the multi-modal learning. Different from these approaches, in this paper, we conduct the RGB-IR object detection task from mono-modality feature learning perspective and propose a novel M$\rm ^2$D-LIF framework that only introduces a lightweight feature fusion module to achieve superior object detection performance.

\subsection{Knowledge Distillation for Object Detection}

Knowledge Distillation (KD)~\cite{hinton2015distilling} can transfer knowledge from a high-capacity teacher model to a compact student model, thereby reducing the complexity of the model while ensuring model performance. For the object detection task, FRS~\cite{zhixing2021distilling} and PGD~\cite{yang2022prediction} leveraged foreground masks to enhance the ability of the student model to capture critical object features. CWD~\cite{shu2021channel} normalized activation maps of each channel to obtain soft probability maps and minimized the KL divergence between these maps, enabling the student to focus on the most salient regions of each channel. Besides, PKD~\cite{cao2022pkd} imitated features with Pearson correlation coefficient to focus on the relational information from the teacher and relax constraints on the magnitude of the features.
Recently, CrossKD~\cite{wang2024crosskd} introduced cross-head knowledge distillation by transferring the intermediate features from the student detection head to the teacher providing a diversified perspective.
As for the multi-modal object detection task, KD is usually used to transfer the knowledge from multi-modality features to mono-modality features. For example, distillation from Radar-Lidar to Radar~\cite{xu2024sckd}, BEV-Lidar to BEV~\cite{hao2024mapdistill}, and RGB-Infrared to RGB~\cite{liu2021deep}.
In our paper, unlike the above methods, we design a novel knowledge distillation method called Mono-Modality Distillation (M$^2$D) to improve the feature representation of each backbone network in multi-modality joint training.

\begin{figure}[t!]
  \centering
      \includegraphics[width=0.9\linewidth]{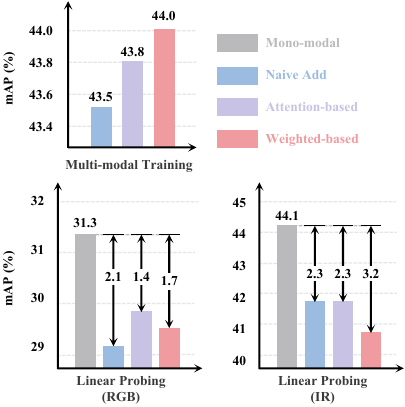}
\vspace{-0.3cm}
  \caption{Liner probing evaluation on the FLIR dataset. Three types of feature fusion methods are selected for comparison, such as naive addition (Halfway Fusion \cite{liu2016multispectral}), weighted-based (IWM~\cite{hu2025ei}), and attention-based (CMX~\cite{zhang2023cmx}).}
  \label{fig:linear_probing}
  \vspace{-0.5cm}
\end{figure}

\section{Methodology}
\subsection{Linear Probing Evaluation} \label{subsec: 3.1}
In the current MMOD framework, each modality is encoded by its corresponding backbone network, and then a fusion module is utilized to obtain the fused features for downstream perception tasks. Although achieving superior performance, we claim that this way will lead to insufficient learning of each modality and thus fail to achieve optimal detection performance. To validate this point, we employ the linear probing evaluation. Specifically, we first train the mono-modal object detectors and the multi-modal object detectors, respectively. For multi-modal object detectors, we select three popular feature fusion methods for comparison, such as Halfway Fusion \cite{liu2016multispectral} (naive addition), IWM~\cite{hu2025ei} (weighted-based), and CMX~\cite{zhang2023cmx} (attention-based). Note that, all detectors utilize the CSPDarknet53~\cite{yolov8_ultralytics} as the backbone network.
After that, each backbone from both the mono- and multi-modality detectors is frozen and connected with the new detection head for training and testing.
As shown in Figure~\ref{fig:linear_probing}, the evaluation results reveal that:

\textbf{(1) The backbone networks from multi-modality joint training are insufficient learning.} 
After linear probing evaluation, we can observe that all backbone networks from multi-modality joint training are worse than those from mono-modality training. This indicates that due to the existence of the fusion module, the learning ability of each backbone network is limited during multi-modal training.

\textbf{(2) The weighted-based method can serve as a competitive fusion way to improve performance.} 
Compared with other fusion methods, the detection performance of the weighted-based backbone network decrease significantly.
This indicates that the weighted-based fusion can achieve higher performance on the weaker backbone networks, which proves its effectiveness in the feature fusion.

\begin{figure*}[!t]
  \centering
      \includegraphics[width=0.95\linewidth]{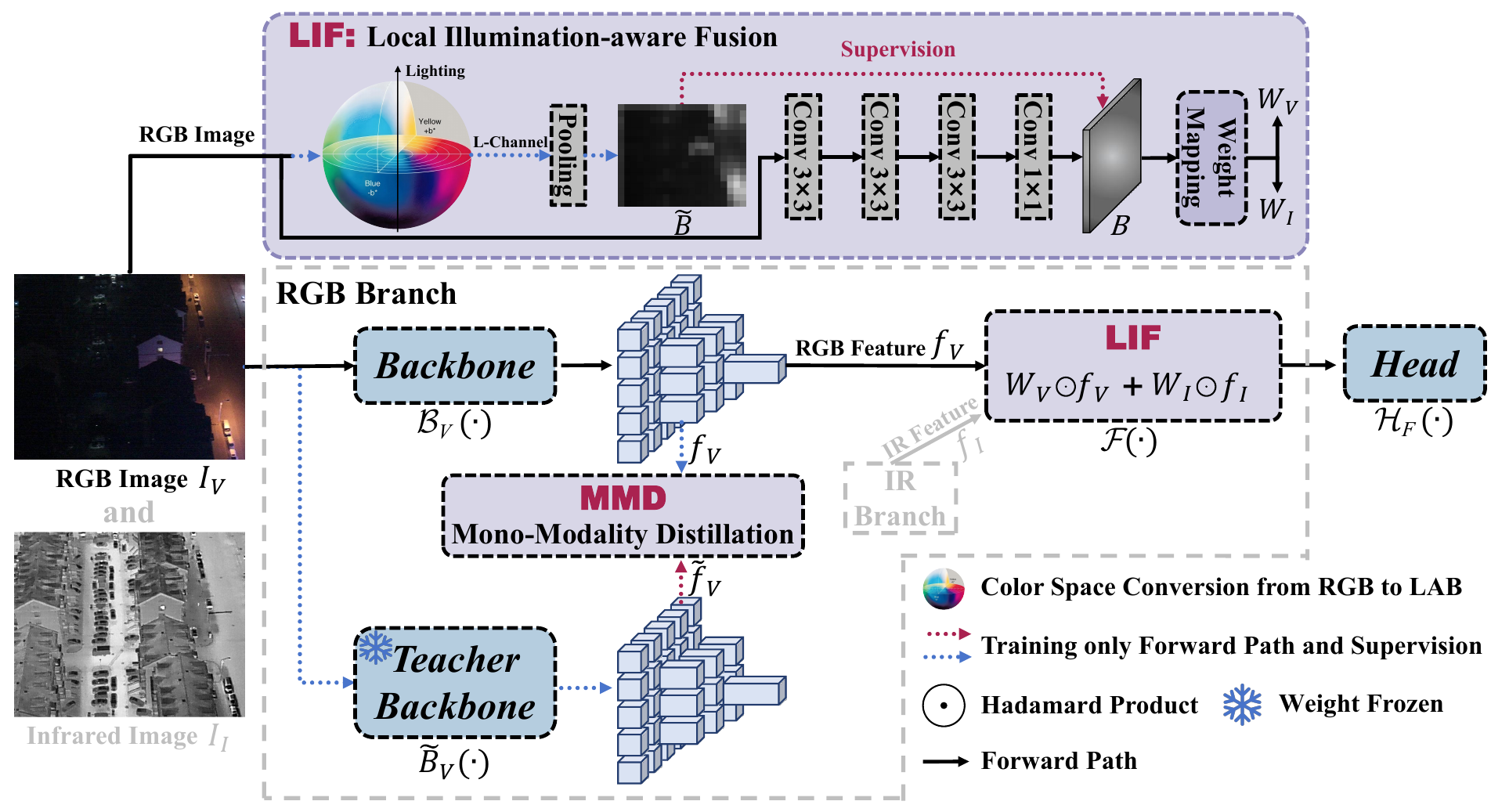}
    \vspace{-0.4cm}
  \caption{Overview of our {$\rm M^2$}D-LIF framework. The proposed M$^2$D is for enhancing mono-modal feature extraction capability of the backbone in the multi-modal object detection task. The proposed LIF method can evaluate the mono-modal quality based on RGB image's local illumination conditions. Noted that since our framework is designed from the mono-modality perspective, the design of the infrared branch and the RGB branch is symmetrical and has the same network structure. Therefore, only the image processing flow of the RGB modality is emphasized in the figure, and the infrared branch is omitted. }

  \label{fig:3}
  \vspace{-0.3cm}
\end{figure*}

\subsection{M$^2$D-LIF Framework}

From Section~\ref{subsec: 3.1}, we have observed that the recent MMOD methods still suffer from the insufficient learning of mono-modality features. To solve this issue, referring to the knowledge distillation technology, we consider utilizing the additional encoder pretrained on the mono-modality as the teacher model to distill the multi-modality backbone network during training. Thus, under the premise that the multi-modal encoder learning sufficiently, we design a novel weighted-based feature fusion method to further improve the performance.
The overall framework (M$^2$D-LIF) is illustrated in Figure~\ref{fig:3}, which mainly consists of \ding{202}~\textbf{Mono-Modality Distillation} and \ding{203}~\textbf{Local Illumination-aware Fusion}. Note that the IR branch is omitted in Figure \ref{fig:3} since it has the same network structure as the RGB branch. 

\ding{202} \textbf{Mono-Modality Distillation (M$^2$D).} 
The M$^2$D aims to enhance the feature extraction capability of the multi-modal encoders, laying a solid foundation for improving object detection performance. 
To bridge the capability gap between the multi- and mono-modal encoders, we first employ a pretrained teacher model of the same modality to distill the multi-modal backbone network using the inner-modality distillation loss $\mathcal{L}_{\text {IM}}$.
To further enhance object-relevant feature extraction, we propose a cross-modality distillation loss $\mathcal{L}_{\text {CM}}$, which leverages cross-modality salient object location priors to guide feature distillation.

Specifically, for the paired RGB image ${I}_V$ and IR image ${I}_I$, we first input them into the teacher and student backbone networks simultaneously, where $\mathcal{B}_V$ and $\mathcal{B}_I$ are the student backbone networks for multi-modal object detection, while $\widetilde{\mathcal{B}}_V$ and $\widetilde{\mathcal{B}}_I$ are the mono-modal teacher backbone networks. Therefore, the output features extracted by each backbone network can be expressed as:
\begin{equation}
\begin{aligned}
        \text{Student:}\quad \boldsymbol f_V = \mathcal{B}_V({I}_V),\quad \boldsymbol f_I = \mathcal{B}_I({I}_I),\\
        \text{Teacher:}\quad\widetilde{\boldsymbol f}_V = \widetilde{\mathcal{B}}_V({I}_V),\quad \widetilde{\boldsymbol f}_I = \widetilde{\mathcal{B}}_I({I}_I),
\end{aligned}
\end{equation}
where $f_I$ and $f_V$ are the outputs of the student backbones. The $\widetilde{f}_V$ and $\widetilde{f}_I$ are the outputs of the teacher backbones. Based on these features, we design the Mono-modality Distillation loss to optimize the multi-modal backbone network. The Mono-modality Distillation loss consists of two components: the inner-modality feature distillation loss~$\mathcal{L}_{\text {IM}}$ and the cross-modality feature distillation loss $\mathcal{L}_{\text {CM}}$.

The loss function $\mathcal{L}_{\text{IM}}$ represents the inner-modality feature distillation loss, which guides the backbone of the multi-modal object detection model to learn intermediate layer responses from a teacher backbone of the same modality.
By minimizing $\mathcal{L}_{\text{IM}}$, the multi-modal backbone is encouraged to align with the feature responses of the teacher model. The loss is defined as follows:
\begin{equation}
\mathcal{L}_{\text{IM}}= \text{D}(f_V,\widetilde{f}_V) + \text{D}(f_I,\widetilde{f}_I), 
\label{eq:lim}
\end{equation}
where $D(\cdot,\cdot)$ denotes the specific distillation method. 

As for the loss function $\mathcal{L}_{\text{CM}}$, it represents the cross-modality feature distillation loss, which integrates cross-modality target location priors.
Specifically, in MMOD tasks, different modalities generally contain information about the same objects. We first employ an attention mechanism to extract the salient object feature attention map, which serves as the location prior. The attention map is then used as a mask to guide object-relevant feature learning. To maintain a lightweight design, we adopt the parameter-free attention method SimAM~\cite{yang2021simam}. The attention map $\widetilde{\mathcal{M}}$ can be calculated as: 
\begin{equation}
\widetilde{\mathcal{M}} =  \text{Sigmoid}(\frac{(\widetilde{f}-\widetilde{\mu})^2+2\widetilde{\sigma} ^2 +2\lambda}{4(\widetilde{\sigma}^2 + \lambda)}),
\end{equation}
where $\widetilde{\mu}$ represents the mean of  $\widetilde{f}$ across spatial dimensions, $\widetilde{\sigma}^2$ represents the variance of $\widetilde{f}$, and $\lambda$ is a small positive constant added for numerical stability. The cross-modality feature distillation loss $\mathcal{L}_{\text{CM}}$ is formulated as follows:
\begin{equation}
\begin{aligned}
\mathcal{L}_{\text{CM}}&= \text{D}(\widetilde{\mathcal{M}}_V\odot f_I,\widetilde{\mathcal{M}}_V\odot\widetilde{f}_V) +\\ &\text{D}(\widetilde{\mathcal{M}}_I\odot f_V,\widetilde{\mathcal{M}}_I\odot\widetilde{f}_I), \\
\end{aligned}
\label{eq:lcm}
\end{equation}
where $\widetilde{\mathcal{M}}_V$ and $\widetilde{\mathcal{M}}_I$ are the attention maps of different modalities. The overall loss function of M$^2$D is defined as the sum of the inner- and cross-modality loss:
\begin{equation}
\mathcal{L}_{M^2D}= \mathcal{L}_{\text {IM}}+\mathcal{L}_{\text {CM}}.
\end{equation}

\ding{203} \textbf{Local Illumination-aware Fusion (LIF).}
After ensuring sufficient learning of mono-modality features, we design a weighted-based fusion method called Local Illumination-aware Fusion (LIF) to explicitly achieve complementary fusion. Different from previous weighted-based methods that only provide one weight for the entire RGB image, the LIF module provides a weight map through the brightness prediction, which can dynamically set different weights for different illumination region features.
Specifically, as illustrated in Figure~\ref{fig:3}, the LIF module is constructed with several convolutions and one activation layer, which can be formulated as:
 \begin{equation}
     B = ConvBlock(I_V),
 \end{equation}
where $B$ denotes the predicted brightness map. We transform the RGB image to the $LAB$ color space and extract the $L$ channel as the ground-truth $\widetilde{B}$ to supervise the brightness prediction, which is formulated as follows:
\begin{equation}
    \mathcal{L}_{LI} = ||B, \widetilde{B}||_2.
\end{equation}

Based on the predicted brightness map $B$, we design the following weight generation mechanism to adaptively adjust the weights matrix of different modality features, which is calculated as follows:

\begin{equation}\label{eq:weight}
\left\{ \begin{aligned}
	&W_V=\beta \times \min\mathrm{(}\frac{B-\alpha}{2\alpha},\frac{1}{2})+\frac{1}{2},\\
	&W_I=1-W_V,\\
\end{aligned}\,\, \right. 
\end{equation} 
where $W_V$ and $W_I$ represent the weight of the RGB and infrared modalities, respectively. The hyperparameter $\alpha$ is the threshold that determines the importance of RGB features, and $\beta$ is the amplitude of the $W_{V}$. Thus, the boundary of $W_{V}$ and $W_{I}$ are:
\begin{equation}
    \frac{1-\beta}{2} \leq  W_{V},  W_{I}   \leq \frac{1+\beta}{2}.
\end{equation}

Finally, the LIF module uses the weight maps $W_{V}$ and $W_{I}$ to perform element-wise weighted fusion for the multi-modal features. Therefore, the final fused feature can be represented as:
\begin{equation}
   f^i_F = \mathcal{F}(f_{V},f_{I}) =  W^i_{V}\odot f^i_{V}+W^i_{I}\odot f^i_{I}.
\end{equation}

 \ding{204} \textbf{Training and Inference.}
For the training stage, we utilize $\mathcal{L}_{M^2D}$ and $\mathcal{L}_{LI}$ to optimize our M$^2$D-LIF framework. Therefore, for the MMOD task, the overall loss function is formulated as follows:
\begin{equation}
\label{eq:overall}
\begin{aligned}
\mathcal{L} & =\mathcal{L}_{\mathrm{det}} + \lambda_{M^2D} \mathcal{L}_{M^2D}+\lambda_{LI}\mathcal{L}_{LI},
\end{aligned}
\end{equation}
where $\lambda_{M^2D}$ and $\lambda_{LI}$ are the hyperparameters that control the balance between each loss.

As for the inference, we only utilize the multi-modality backbone network with our proposed LIF module to perform the multi-modal object detection task.

\begin{table}[!t]
\centering
    \setlength{\tabcolsep}{1mm}
        \renewcommand{\arraystretch}{1.0}
    \caption{Comparison of detection performance (mAP, in\%) and computational cost (Params, FLOPs) between the LIF and other SOTA fusion methods on the FLIR (F), DroneVehicle (D), and LLVIP (L) datasets. The symbol $^\star$ indicates training with our M$^2$D method. The Best results are highlighted by \textbf{bold}.}
    \vspace{-0.3cm}
    \label{tab:1}
\begin{tabular}{@{}c|c|ccc|cc@{}}
\toprule
\begin{tabular}[c]{@{}c@{}}Fusion\\ Method\end{tabular}&\begin{tabular}[c]{@{}c@{}}Fusion\\ Type\end{tabular} & F & D & L & \begin{tabular}[c]{@{}c@{}}Params\\ (M)\end{tabular}                   & \begin{tabular}[c]{@{}c@{}}Flops\\ (G)\end{tabular}\\ \midrule

Naive add &-&  43.5 & 66.9 & 68.2&36.47&116.7\\ \midrule

CMX~\cite{zhang2023cmx} &\multirow{2}{*}{Attention}                                                  & 43.8 & 68.0                                                    &    67.3  & \multirow{2}{*}{+22.3} & \multirow{2}{*}{+33.7}   \\

$^\star$CMX~\cite{zhang2023cmx}                                              && 44.1 & 68.9                                                    &   69.5   &                          &                           \\
\midrule
IWM \cite{hu2025ei}                                                 &\multirow{2}{*}{Weighted}    & 44.0 & 68.9                                                        & 68.8      & \multirow{2}{*}{+1.1}   & \multirow{2}{*}{+20.1}   \\
$^\star$IWM \cite{hu2025ei}                                         &   & 44.9 &  70.4                                                     &       69.3    &                          &                           \\ \midrule
LIF (Ours)         & \multirow{2}{*}{Weighted} &   44.9   &            68.3                                             &   67.9    & \multirow{2}{*}{\textbf{+0.06}}  & \multirow{2}{*}{\textbf{+6.5}}    \\
$^\star$LIF  (Ours)   &                                      &  \textbf{46.1}    &   \textbf{70.6}                                                  &  \textbf{70.8}    &                          &                           \\ 

\bottomrule
\end{tabular}
    \vspace{-0.3cm}
\end{table}

\subsection{Why Using LIF Module for Feature Fusion}
Actually, various fusion methods can be utilized with our proposed M$^2$D methods to further improve object detection performance, as shown in Table~\ref{tab:1}. However, we introduce the LIF module into our framework to cooperate with the M$^2$D method for the following two reasons:

\textbf{(1) robustness-and-reasonable.}
As analyzed in Section~\ref{subsec: 3.1}, the weighted-based method can serve as a competitive fusion way to improve performance. As a weighted-based fusion module, LIF is more robust to the insufficient learning of mono-modality encoders, thus achieving superior performance. Furthermore, unlike current weighted-based methods, the LIF module provides reasonably fine-grained quality assessment for the mono-modality images and explicitly achieves the complementary fusion.

 \textbf{(2) lightweight-yet-effective.} 
As shown in Table~\ref{tab:1}, although attention-based methods can adaptively calculate the required features between modalities, they typically rely on high computational complexity. Unlike other methods, our LIF introduces only an additional 0.1M parameters and 6.4 GFLOPs in computational cost to achieve the superior performance, which is more lightweight and effective.

\label{sec:experiments}

\section{Experiments}
\subsection{Datasets and Evaluation Metrics}

\noindent\textit{ 1) \textbf{DroneVehicle:}} This dataset~\cite{sun2022drone} includes images captured by drones in urban areas under different lighting conditions.  It has annotations with oriented bounding boxes for five categories: `car`, `truck', `bus', `van', and `freight car'. There are 28,439 pairs of RGB and IR images, with 17,990 for training, 1,469 for validation, and 8,980 for testing.

\noindent\textit{ 2) \textbf{FLIR-aligned:}} This dataset~\cite{9191080} contains paired RGB and IR images in day and night scenes. It includes 5,142 aligned RGB-IR pairs, with 4,129 for training and 1,013 for testing, focusing on `person', `car', and `bicycle' categories. The `dog' category was removed due to its rarity.

\noindent\textit{ 3) \textbf{LLVIP:}} This dataset ~\cite{jia2021llvip} is designed for low-light environments, featuring 15,488 strictly aligned RGB-IR image pairs, mostly in very dark scenes. It is divided into 12,025 pairs for training and 3,463 pairs for testing.

\begin{table}[!t]
\caption{Ablation on our M$^2$D-LIF framework with performance measured by mAP (\%). The best results are highlighted in \textbf{bold}. Note: Params/FLOPs tested on FLIR during reference.}
\vspace{-0.2cm}
\label{tab:ablationStudy}
    \setlength{\tabcolsep}{0.9mm}
        \renewcommand{\arraystretch}{1.0}
\centering
\begin{tabular}{@{}cc|ccc|cc@{}}
\toprule
M$^2$D    &LIF  & FLIR    & DroneVehicle    & LLVIP &Params&FLOPs   \\ \midrule
               &     & 43.5 & 66.9 & 68.2 &\multirow{2}{*}{36.47M}&\multirow{2}{*}{116.7G}\\

   \ding{51}   &       &   45.1   &  69.2    &   70.7 && \\\cline{6-7}
   & \ding{51}& 45.0 & 68.5 & 69.4&&\\
\ding{51} &     \ding{51}       &   \textbf{46.1}   &  \textbf{70.6}    &  \textbf{70.8}   &\multirow{-2}{*}{36.53M}&\multirow{-2}{*}{123.2G} \\ \bottomrule
\end{tabular}
\vspace{-0.1cm}
\end{table}

\begin{table}[!t]
\caption{Ablation on different ways of the weight map generation in the LIF module with performance measured by mAP (\%). The best results are highlighted in \textbf{bold}. Sup. represents supervision.}
\vspace{-0.2cm}
\label{tab:LIF}
    \setlength{\tabcolsep}{2.3mm}
        \renewcommand{\arraystretch}{1.0}
\centering
\begin{tabular}{@{}c|ccc@{}}
\toprule
Fusion weight & FLIR    & DroneVehicle    & LLVIP    \\ \midrule
L-channel  & 44.2 & 67.4 & 70.2 \\
w/o L-channel Sup. & 45.0  &    69.8  & 70.3 \\
w/ L-channel Sup. & \textbf{46.1}  &    \textbf{70.6}  & \textbf{70.8}    \\ \bottomrule

\end{tabular}
\vspace{-0.3cm}
\end{table}

\noindent\textit{ 4) \textbf{Mean Average Precision (mAP):}} mAP is a standard metric for object detection that evaluates performance based on classification accuracy and Intersection over Union (IoU). Specifically, mAP$_{50}$ denotes the mAP across all classes at a fixed IoU threshold of 0.50. The mAP is the average of mAP values calculated over a range of IoU thresholds from 0.50 to 0.95 with a step of 0.05.

\subsection{Implementation Details}


All experiments are conducted on NVIDIA GeForce RTX 4090 GPUs. We employ CSPDarknet53 as the backbone network and modified it to process dual-modal inputs. The network is optimized using SGD with a momentum of $0.937$ and a weight decay of $5\times 10^{-4}$. We initialize the learning rate at $1\times 10^{-2}$ and gradually decrease it to $1\times 10^{-4}$ throughout the training process. In the testing phase, we set the confidence threshold to $0.25$ to filter detection results. If there is no special state, the ablation study results on the DroneVehicle dataset are conducted on the validation set.

\subsection{Ablation Study}

\noindent\textit{ 1) \textbf{Study on Each Component:}}
To evaluate the effectiveness of the M$^2$D-LIF framework, we conducted ablation studies on the M$^2$D and LIF modules. As shown in Table~\ref{tab:ablationStudy}, we incrementally applied these components to the baseline model to evaluate their individual contributions. Without either of these two modules, the model achieves mAP scores of 44.4\%, 66.9\%, and 68.2\% on the FLIR, DroneVehicle, and LLVIP datasets, respectively. The integration of the M$^2$D training improves model performance by 1.6\% in FLIR, 2.3\% in DroneVehicle, and 2.5\% in LLVIP and LIF fusion module improves by 1.5\%, 1.6\%, and 1.2\%, respectively.
The best results are achieved when combining both the M$^2$D and LIF modules, yielding significant mAP increases of 2.6\%, 3.7\%, and 2.6\% over the baseline while requiring only a marginal 0.1\% parameter increase and 5.6\% additional computational overhead.
These improvements demonstrate the significant impact of our complete M$^2$D-LIF framework. 
\begin{figure}[t]
  \centering
    \includegraphics[width=\linewidth]{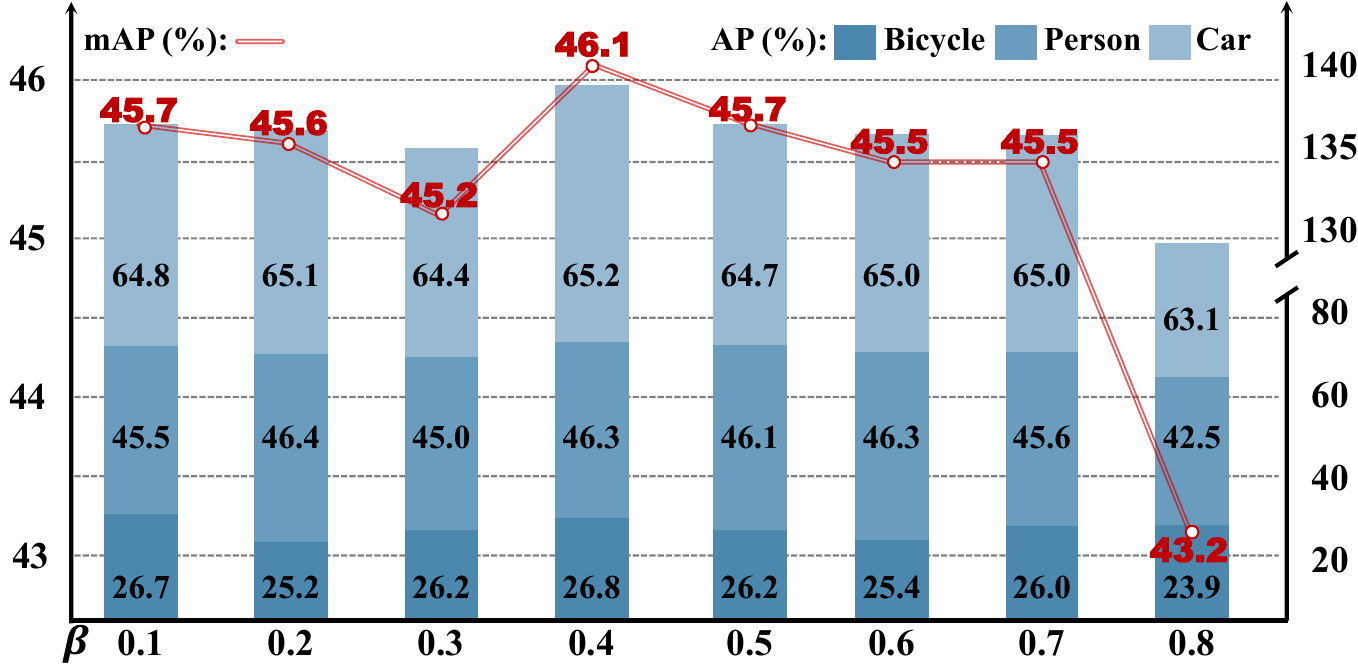} 
  \caption{Ablation on the hyper-parameter $\beta$ in the Equation \ref{eq:weight}.}
    \label{fig:4}
\vspace{-0.3cm}
\end{figure}

\begin{figure*}[!t]
  \centering
      \includegraphics[width=\linewidth]{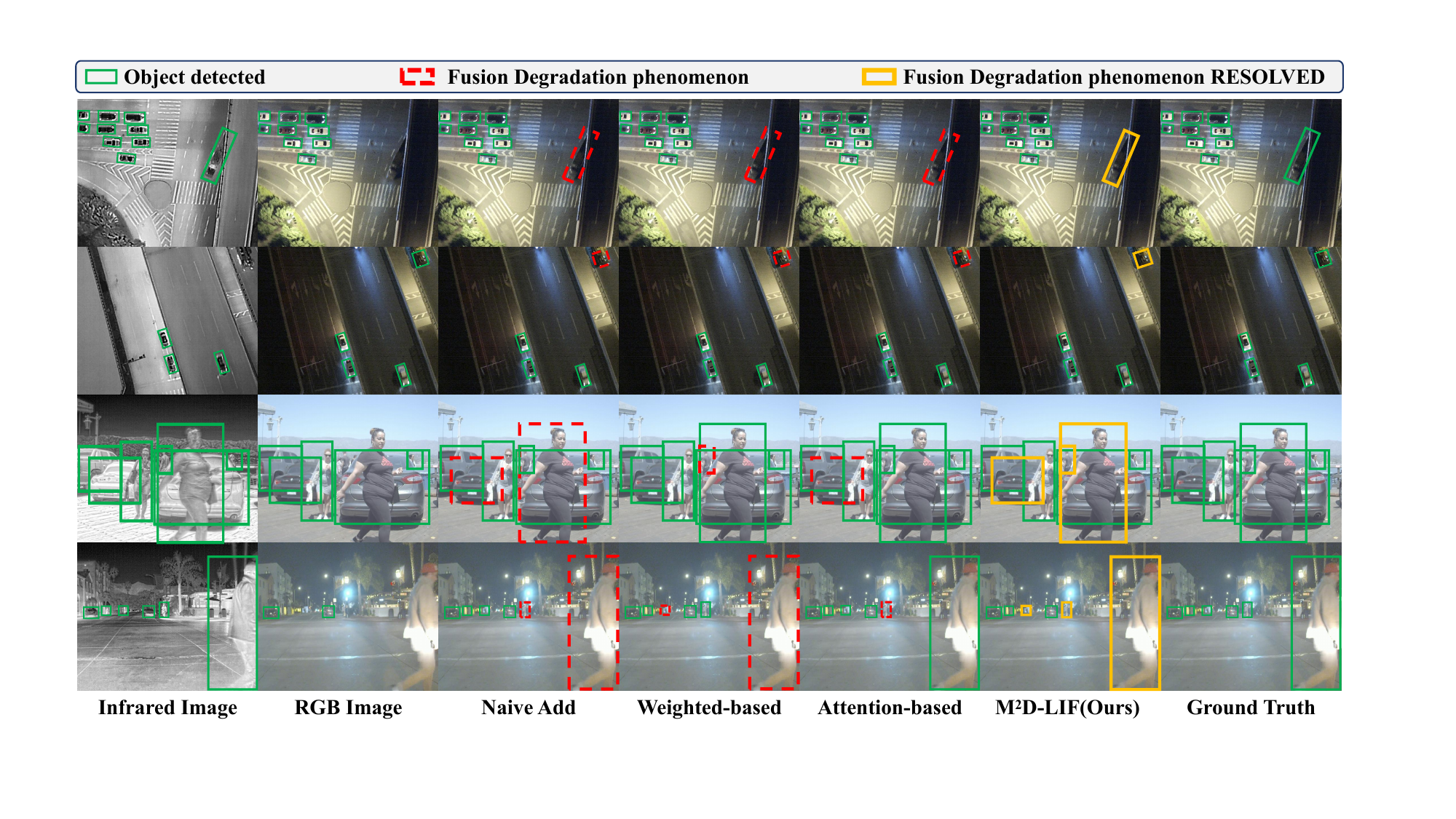}
  \caption{Visualization of the Fusion Degradation phenomenon across the DroneVehicle (first and second rows) and FLIR (third and fourth rows) datasets. Columns from left to right represent infrared modality, RGB modality, naive add method, weighted-based method, attention-based method, our M$^2$D-LIF, and the ground truth. The \textcolor[rgb]{0, 0.6875, 0.3125}{Green boxes} indicate correctly detected objects, while the \textcolor[rgb]{1, 0, 0}{red dashed boxes} represent the Fusion Degradation phenomenon, where objects are detected by mono-modal models but missed by the multi-modal object detection model. The \textcolor[rgb]{1, 0.75, 0}{Orange boxes} emphasize how our proposed method effectively resolves this issue. Zoom in for details.}
  \vspace{-0.3cm}
  \label{fig:5}
\end{figure*}

\begin{figure}[t]
  \centering
    \includegraphics[width=\linewidth]{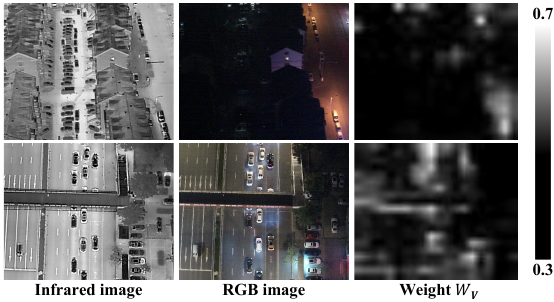} 
   \caption{Visualization of the Weight Map $W_V$ in Equation~\ref{eq:weight}. From left to right are the infrared image, the RGB image, the weight map, and the grayscale color corresponding to the value.}
    \label{fig:6}
    \vspace{-0.5cm}
\end{figure}

\noindent \textit{ 2) \textbf{Study on the LIF:}} To demonstrate that our proposed LIF module surpasses simply sensing illumination, we conducted a comparative experiment against the method directly using the normalized L-Channel from LAB color space as fusion weight. As shown in Table~\ref{tab:LIF}, our LIF module outperforms this approach by 1.9\%, 3.2\%, and 0.6\% on the FLIR, DroneVehicle, and LLVIP datasets, respectively. 
Simply using the normalized L-Channel as the fusion weight cannot enable dynamic learning of the fusion weight, which is not conducive to the complementary fusion of multiple modalities. In the FLIR dataset, it even underperforms compared to the baseline method.
These results further confirm that our LIF module can not only utilize the illumination context but also effectively leverage the complementary relationship between different modalities.

\noindent\textit{ 3) \textbf{Study on the Hyper-parameter $\beta$:}} To investigate the impact of the $\beta$ value in the Equation \ref{eq:weight} on M$^2$D-LIF framework performance, we conducted ablation studies across a pre-defined set of $\beta$ values. As shown in Figure~\ref{fig:4}, 
The bar chart represents the AP of different classes, and the line chart represents the mAP. When beta takes the value of 0.4, our method achieves the best result.
If the value of beta exceeds 0.8, it may lead to the modality imbalance, thus causing a serious decline in performance.
More ablation studies are detailed in the supplementary material.

\subsection{Visualization}

\noindent\textit{ 1) \textbf{Visualization of the Fusion Degradation phenomenon.}}
To demonstrate that our method can effectively mitigate the Fusion Degradation phenomenon, as illustrated in Figure~\ref{fig:5}, we visualized the detection results from the mono-modal method, naive addition (Halfway Fusion \cite{liu2016multispectral}), weighted-based (IWM \cite{hu2025ei}), attention-based (CMX \cite{zhang2023cmx}) method and our proposed M$^2$D-LIF framework on the DroneVehicle (first and second rows) and FLIR dataset (third and fourth rows). 
These examples highlight the widespread occurrence of the Fusion Degradation phenomenon. In contrast, our proposed M$^2$D-LIF framework effectively addresses this issue, ensuring that objects detectable by mono-modal methods remain detectable when using the multi-modal approach, thereby enhancing detection performance.

\noindent\textit{ 2) \textbf{Visualization of the LIF Weight Map:}}
To evaluate whether our LIF module effectively perceives illumination and dynamically assigns different weights to regions with varying lighting conditions, we visualize the weight map $W_V$ in Equation~\ref{eq:weight}. As shown in Figure~\ref{fig:6}, object regions with high-quality illumination in RGB images are assigned with higher weights, thereby explicitly achieving complementary feature fusion. More visualization results are detailed in the supplementary material.

\begin{table*}[!t]
\begin{center}
    \caption{Camparison of the performance measured by mAP$_{50}$, mAP on the DroneVehicle dataset. The best results are highlighted in \underline{\textcolor{red}{\textbf{red}}} and the second-place results are highlighted in \textcolor{blue!70}{\textbf{blue}}. Noted that it uses the `OBB' detectors.}
      \vspace{-0.2cm}
    \label{tab:dronevehicle_compare}
    \setlength{\tabcolsep}{1.10mm}
        \renewcommand{\arraystretch}{1.0}
\begin{tabular}{@{}l|c|ccccccc|cccccc|c@{}}
\toprule
\multirow{2}{*}{\textbf{Method}} &
  \multirow{2}{*}{\textbf{Modality}} &
  \multicolumn{7}{c|}{\textbf{DroneVehicle test}} &
  \multicolumn{6}{c|}{\textbf{DroneVehicle val}} &
  \multirow{2}{*}{\textbf{Params}} \\ 
 &
   &
  \textbf{Car} &
  \textbf{Tru} &
  \textbf{Fre} &
  \textbf{Bus} &
  \textbf{Van} &
  \textbf{mAP$_{50}$} &
  \textbf{mAP} &
  \textbf{Car} &
  \textbf{Tru} &
  \textbf{Fre} &
  \textbf{Bus} &
  \textbf{Van} &
  \textbf{mAP$_{50}$} &
   \\ \midrule
RetinaNet \cite{lin2017focal}      & \multirow{4}{*}{RGB} & 67.5 & 28.2 & 13.7 & 62.1 & 19.3 & 38.1 & 23.4 & 78.8 & 39.9 & 19.5 & 67.3 & 24.9 & 46.1 &     36.4M   \\
Faster R-CNN \cite{ren2017faster}   &                      & 67.9 & 38.6 & 26.3 & 67.0 & 23.2 & 44.6 & 28.4 & 79.0 & 49.0   & 37.2 & 77.0 & 37.0 & 55.9 &  41.8M      \\
S$^2$A Net \cite{han2021align}     &                      & 88.6 & 58.7 & 37.3 & 85.2 & 41.4 & 62.2 & 36.9 & 80.0 & 54.2 & 42.2 & 84.9 & 43.8 & 61.0 &   33.3M     \\
YOLOv8m \cite{yolov8_ultralytics}   &                      & 91.0 & 55.8 & 43.1 & 86.3 & 43.4 & 63.9 & 45.9 & 91.3 & 56.3 & 44.1 & 87.7 & 46.7 & 65.2 &   26.4M   \\ \midrule
RetinaNet \cite{lin2017focal}      & \multirow{4}{*}{IR}  & 79.9 & 32.8 & 28.1 & 67.3 & 16.4 & 44.9 & 27.8 & 89.3 & 38.2 & 40.0 & 79.0 & 32.1 & 55.7 &   36.4M     \\
Faster R-CNN \cite{ren2017faster}   &                      & 88.6 & 42.5 & 35.2 & 77.9 & 28.5 & 54.6 & 31.1 & 89.4 & 53.5 & 48.3 & 87.0 & 42.6 & 64.2 &     41.8M   \\
S$^2$A Net \cite{han2021align}     &                      & 90.0 & 58.3 & 42.9 & 87.5 & 38.9 & 63.5 & 38.7 & 89.9 & 54.5 & 55.8 & 88.9 & 48.4 & 67.5 &     33.3M   \\
YOLOv8m \cite{yolov8_ultralytics}   &                      & 96.5 & 63.3 & 54.6 & 90.6 & 45.6 & 70.1 & 53.7 & 96.4 & 57.8 & 57.7 & 92.7 & 53.2 & 71.6 &     26.4M    \\ \midrule

TarDAL \cite{liu2022target}         &\multirow{9}{*}{RGB+IR}                      & 89.5 & 68.3 & 56.1 & 89.4 & 59.3 & 72.6 & \textcolor{blue!70}{\textbf{43.3}} & -    & -    & -    & -    & -    & -    &    -    \\
UA-CMDet \cite{sun2022drone}       &                      & 87.5 & 60.7 & 46.8 & 87.1 & 38.0 & 64.0 & 40.1 & -    & -    & -    & -    & -    & -    &    -    \\
GLFNet \cite{kang2024global}       &                      & \textcolor{blue!70}{\textbf{90.3}} & 72.7 & 53.6 & 88.0 & 52.6 & 71.4 & 42.9 & -    & -    & -    & -    & -    & -    &      -  \\

TSFADet \cite{yuan2022translation} &                      & 89.2 & 72.0 & 54.2 & 88.1 & 48.8 & 70.4 & -    & 89.9 & 67.9 & 63.7 & 89.8 & 54.0 & 73.1 &      104.7M  \\
CALNet~\cite{he2023multispectral} &                      & \textcolor{blue!70}{\textbf{90.3}} & 76.2 & 63.0 & 89.1 & 58.5 & 75.4 & -    & 90.3 & \textcolor{blue!70}{\textbf{73.7}} & \textcolor{blue!70}{\textbf{68.7}} & 89.7 & \textcolor{blue!70}{\textbf{59.7}} & \textcolor{blue!70}{\textbf{76.4}} &    -    \\
C$^2$Former \cite{yuan2024c}       &                      & 90.0 & 72.1 & 57.6 & 88.7 & 55.4 & 72.8 & 42.8    & 90.2 & 68.3 & 64.4 & 89.8 & 58.5 & 74.2 &   \textcolor{blue!70}{\textbf{100.8M}}     \\
CAGTDet \cite{yuan2024improving} & & - & - & - & - & - & - & -& \textcolor{blue!70}{\textbf{90.8}} & 69.7 & 66.3& \textcolor{blue!70}{\textbf{90.5}} & 55.6 & 74.6 & 140.3M \\
OAFA~\cite{chen2024weakly} & & \textcolor{blue!70}{\textbf{90.3}} & \textcolor{blue!70}{\textbf{76.8}} & \underline{\textcolor{red}{\textbf{73.3}}} & \textcolor{blue!70}{\textbf{90.3}} & \underline{\textcolor{red}{\textbf{66.0}}} &\textcolor{blue!70}{\textbf{79.4}} & -  & -  & -  & -  & -  & -  & -   & -  \\
\textbf{M$^2$D-LIF (Ours)} &
   &
  \underline{\textcolor{red}{\textbf{97.8}}} &
  \underline{\textcolor{red}{\textbf{81.0}}} &
  \textcolor{blue!70}{\textbf{67.9}} &
 \underline{\textcolor{red}{\textbf{96.0}}} &
  \textcolor{blue!70}{\textbf{64.6}} &
  \underline{\textcolor{red}{\textbf{81.4}}} &
  \underline{\textcolor{red}{\textbf{68.1}}} &
  
  \underline{\textcolor{red}{\textbf{98.1}}} &
  \underline{\textcolor{red}{\textbf{81.6}}} &
 \underline{\textcolor{red}{\textbf{76.5}}} &
 \underline{\textcolor{red}{\textbf{96.4}}} &
  \underline{\textcolor{red}{\textbf{69.7}}} &
  \underline{\textcolor{red}{\textbf{{84.5}}}} &
  \underline{\textcolor{red}{\textbf{37.1M}}} \\ \bottomrule

\end{tabular}
  \vspace{-0.7cm}
\end{center}
\end{table*}

\begin{table}[!t]
\begin{center}
    \caption{Camparison of the performance measured by mAP (\%) on the FLIR and LLVIP datasets. The best results are highlighted in \underline{\textcolor{red}{\textbf{red}}} and the second-place are highlighted in \textcolor{blue!70}{\textbf{blue}}.}
  \vspace{-0.2cm}
    \label{tab:flir_compare}
    \setlength{\tabcolsep}{1.1mm}
        \renewcommand{\arraystretch}{1.0}
\begin{tabular}{@{}c|c|cc|c@{}}
\toprule
\textbf{Method}                             & \textbf{Modality}       & \textbf{FLIR} & \textbf{LLVIP} & \textbf{Params.} \\ \midrule
RetinaNet \cite{lin2017focal}           &         \multirow{4}{*}{RGB}                 & 21.9          & 42.8           & 35.9M                  \\
Faster R-CNN \cite{ren2017faster}       &                         & 28.9          & 45.1           &           41.3M     \\
DDQ-DETR \cite{zhang2023dense}          &                         & 30.9          & 46.7           & -               \\
YOLOv8m \cite{yolov8_ultralytics}       &                         & 27.8          & 51.7         &25.9M          \\ \midrule
RetinaNet \cite{lin2017focal}           &             \multirow{4}{*}{IR}             & 31.5          & 55.1           & 35.9M               \\
Faster R-CNN \cite{ren2017faster}       &                         & 37.6          & 54.5           &  41.3M            \\
DDQ-DETR \cite{zhang2023dense}          &                         & 37.1          & 58.6           &   -             \\
YOLOv8m \cite{yolov8_ultralytics}       &                         & 36.5          & 58.9           & 25.9M           \\ \midrule
Halfway Fus. \cite{liu2016multispectral} & \multirow{8}{*}{RGB+IR} & 35.8          & 55.1           &     -             \\
GAFF \cite{zhang2021guided}             &                         & 37.3          & 55.8           & \textcolor{red}{\textbf{31.4M }}           \\
CFT \cite{qingyun2021cross}             &                         & 40.2          & 63.9           & 206.0M          \\
CSAA \cite{cao2023multimodal}           &                         & 41.3          & 59.2           & -                \\
ICAFusion \cite{shen2024icafusion}      &                         & 41.4          & 64.3           & 120.2M           \\
RSDet \cite{zhao2024removal}            &                         & 43.8          & 61.3           & 386.0M             \\
UniRGB-IR \cite{yuan2024unirgb}         &                         & \textcolor{blue!70}{\textbf{44.3}}         & \textcolor{blue!70}{\textbf{63.2}}           & 147.0M           \\
\textbf{M$^2$D-LIF (Ours)}                               &                         & \textcolor{red}{\textbf{46.1}} & \textcolor{red}{\textbf{70.8}}  & \textcolor{blue!70}{\textbf{36.5M}}  \\ \bottomrule
\end{tabular}%
  \vspace{-0.5cm}
\end{center}
\end{table}

\subsection{Comparison With State-of-the-Art Methods}

\noindent\textit{ 1) \textbf{On DroneVehicle:} }We evaluated the detection performance and parameter count of our method against four mono-modal oriented bounding box (OBB) detection methods, as well as eight SOTA multi-modal OBB detection methods on the DroneVehicle test and validation set. Table~\ref{tab:dronevehicle_compare} presents the comparative results. Notably, our M$^2$D-LIF framework achieves the highest mAP$_{50}$ and mAP of 81.4\% and 68.1\% on the test set. Additionally, it recorded the highest mAP$_{50}$ of 85.2\% on the validation set. Furthermore, our method attained the highest mAP$_{50}$ across all five categories on the validation set and three first-place and two second-place rankings on the test set. 
Our method has not only achieved state-of-the-art (SOTA) performance, but also the parameter count of our M$^2$D-LIF framework is optimal with 37.1M. It has only increased by 0.1M compared to the baseline model in the test phase, which demonstrates its superiority over other multi-modal methods.

\noindent\textit{ 2) \textbf{On FLIR and LLVIP: }}Furthermore, we conducted comprehensive comparisons of our method against state-of-the-art object detection methods on the FLIR and LLVIP datasets, involving four mono-modal and seven multi-modal methods, as presented in Table~\ref{tab:flir_compare}. Our M$^2$D-LIF framework achieves 46.1\% and 70.8\% of mAP on the FLIR and LLVIP datasets, respectively, while only employing 36.5M parameters, the second lowest among multi-modal methods. The above results underscore the effectiveness of our M$^2$D-LIF framework in handling diverse datasets. The superior performance on all three datasets highlights the adaptability of the proposed framework. Our method not only competes with but often surpasses existing state-of-the-art multi-modal methods while achieving higher computational efficiency with a lower parameters.

\section{Conclusions}
\label{sec:conclusion}
In this paper, we first observed the fusion degradation phenomenon. Through linear probing evaluation, we further identified mono-modality insufficient learning as its underlying cause of this phenomenon. 
To address this issue, we rethought the multi-modal object detection task from the mono-modality learning perspective and constructed an end-to-end M$^2$D-LIF framework. 
The Mono-Modality Distillation (M$^2$D) method was designed to enhance the feature extraction capability of the multi-modal encoders. The Local Illumination-aware Fusion (LIF) module was designed to dynamically set different weights for the regions with different illumination conditions. 
Extensive experimental results demonstrated that M$^2$D-LIF achieved superior performance.
We believe that our framework holds potential for expansion to other multi-modal tasks.

\newpage
\section*{Acknowledgement}
This work was supported by the Fundamental Research Funds for the Central Universities.

{
    \small
    \bibliographystyle{ieeenat_fullname}
    \bibliography{main}

\begin{thebibliography}{47}
\providecommand{\natexlab}[1]{#1}
\providecommand{\url}[1]{\texttt{#1}}
\expandafter\ifx\csname urlstyle\endcsname\relax
  \providecommand{\doi}[1]{doi: #1}\else
  \providecommand{\doi}{doi: \begingroup \urlstyle{rm}\Url}\fi

\bibitem[Cao et~al.(2022)Cao, Zhang, Gao, Cheng, Cheng, and Cheng]{cao2022pkd}
Weihan Cao, Yifan Zhang, Jianfei Gao, Anda Cheng, Ke Cheng, and Jian Cheng.
\newblock Pkd: General distillation framework for object detectors via pearson correlation coefficient.
\newblock \emph{Advances in Neural Information Processing Systems}, 35:\penalty0 15394--15406, 2022.

\bibitem[Cao et~al.(2023)Cao, Bin, Hamari, Blasch, and Liu]{cao2023multimodal}
Yue Cao, Junchi Bin, Jozsef Hamari, Erik Blasch, and Zheng Liu.
\newblock Multimodal object detection by channel switching and spatial attention.
\newblock In \emph{Proceedings of the IEEE/CVF Conference on Computer Vision and Pattern Recognition Workshops}, pages 403--411, 2023.

\bibitem[Chen et~al.(2024)Chen, Qi, Liu, Bin, Fu, Hu, and Zhong]{chen2024weakly}
Chen Chen, Jiahao Qi, Xingyue Liu, Kangcheng Bin, Ruigang Fu, Xikun Hu, and Ping Zhong.
\newblock Weakly misalignment-free adaptive feature alignment for uavs-based multimodal object detection.
\newblock In \emph{Proceedings of the IEEE/CVF Conference on Computer Vision and Pattern Recognition}, pages 26836--26845, 2024.

\bibitem[Guan et~al.(2019)Guan, Cao, Yang, Cao, and Yang]{guan2019fusion}
Dayan Guan, Yanpeng Cao, Jiangxin Yang, Yanlong Cao, and Michael~Ying Yang.
\newblock Fusion of multispectral data through illumination-aware deep neural networks for pedestrian detection.
\newblock \emph{Information Fusion}, 50:\penalty0 148--157, 2019.

\bibitem[Han et~al.(2021)Han, Ding, Li, and Xia]{han2021align}
Jiaming Han, Jian Ding, Jie Li, and Gui-Song Xia.
\newblock Align deep features for oriented object detection.
\newblock \emph{IEEE transactions on geoscience and remote sensing}, 60:\penalty0 1--11, 2021.

\bibitem[Hao et~al.(2024)Hao, Li, Zhang, Li, Yin, Jung, Park, Yoo, Zhao, and Zhang]{hao2024mapdistill}
Xiaoshuai Hao, Ruikai Li, Hui Zhang, Dingzhe Li, Rong Yin, Sangil Jung, Seung-In Park, ByungIn Yoo, Haimei Zhao, and Jing Zhang.
\newblock Mapdistill: Boosting efficient camera-based hd map construction via camera-lidar fusion model distillation.
\newblock \emph{arXiv preprint arXiv:2407.11682}, 2024.

\bibitem[He et~al.(2023)He, Tang, Zou, and Zhang]{he2023multispectral}
Xiao He, Chang Tang, Xin Zou, and Wei Zhang.
\newblock Multispectral object detection via cross-modal conflict-aware learning.
\newblock In \emph{Proceedings of the 31st ACM International Conference on Multimedia}, pages 1465--1474, 2023.

\bibitem[Hinton et~al.(2015)Hinton, Vinyals, and Dean]{hinton2015distilling}
Geoffrey Hinton, Oriol Vinyals, and Jeff Dean.
\newblock Distilling the knowledge in a neural network.
\newblock \emph{arXiv preprint arXiv:1503.02531}, 2015.

\bibitem[Hu et~al.(2025)Hu, He, Li, Zhao, Chen, and Kang]{hu2025ei}
Ke Hu, Yudong He, Yuan Li, Jiayu Zhao, Song Chen, and Yi Kang.
\newblock Ei 2 det: Edge-guided illumination-aware interactive learning for visible-infrared object detection.
\newblock \emph{IEEE Transactions on Circuits and Systems for Video Technology}, 2025.

\bibitem[Jia et~al.(2021)Jia, Zhu, Li, Tang, and Zhou]{jia2021llvip}
Xinyu Jia, Chuang Zhu, Minzhen Li, Wenqi Tang, and Wenli Zhou.
\newblock Llvip: A visible-infrared paired dataset for low-light vision.
\newblock In \emph{Proceedings of the IEEE/CVF international conference on computer vision}, pages 3496--3504, 2021.

\bibitem[Jocher et~al.(2023)Jocher, Chaurasia, and Qiu]{yolov8_ultralytics}
Glenn Jocher, Ayush Chaurasia, and Jing Qiu.
\newblock Ultralytics yolov8, 2023.

\bibitem[Kang et~al.(2024)Kang, Yin, and Duan]{kang2024global}
Xudong Kang, Hui Yin, and Puhong Duan.
\newblock Global--local feature fusion network for visible--infrared vehicle detection.
\newblock \emph{IEEE Geoscience and Remote Sensing Letters}, 21:\penalty0 1--5, 2024.

\bibitem[Li et~al.(2018)Li, Song, Tong, and Tang]{li2018multispectral}
Chengyang Li, Dan Song, Ruofeng Tong, and Min Tang.
\newblock Multispectral pedestrian detection via simultaneous detection and segmentation.
\newblock In \emph{British Machine Vision Conference (BMVC)}, 2018.

\bibitem[Li et~al.(2019)Li, Song, Tong, and Tang]{li2019illumination}
Chengyang Li, Dan Song, Ruofeng Tong, and Min Tang.
\newblock Illumination-aware faster r-cnn for robust multispectral pedestrian detection.
\newblock \emph{Pattern Recognition}, 85:\penalty0 161--171, 2019.

\bibitem[Li et~al.(2024)Li, Wang, Hu, Li, Ni, Zhao, and Wang]{li2024fd2}
Ke Li, Di Wang, Zhangyuan Hu, Shaofeng Li, Weiping Ni, Lin Zhao, and Quan Wang.
\newblock Fd2-net: Frequency-driven feature decomposition network for infrared-visible object detection.
\newblock \emph{arXiv preprint arXiv:2412.09258}, 2024.

\bibitem[Lin et~al.(2017)Lin, Goyal, Girshick, He, and Doll{\'a}r]{lin2017focal}
Tsung-Yi Lin, Priya Goyal, Ross Girshick, Kaiming He, and Piotr Doll{\'a}r.
\newblock Focal loss for dense object detection.
\newblock In \emph{Proceedings of the IEEE international conference on computer vision}, pages 2980--2988, 2017.

\bibitem[Liu et~al.(2016{\natexlab{a}})Liu, Zhang, Wang, and Metaxas]{liu1611multispectral}
J Liu, S Zhang, S Wang, and DN Metaxas.
\newblock Multispectral deep neural networks for pedestrian detection. arxiv 2016.
\newblock \emph{arXiv preprint arXiv:1611.02644}, 2016{\natexlab{a}}.

\bibitem[Liu et~al.(2016{\natexlab{b}})Liu, Zhang, Wang, and Metaxas]{liu2016multispectral}
Jingjing Liu, Shaoting Zhang, Shu Wang, and Dimitris~N Metaxas.
\newblock Multispectral deep neural networks for pedestrian detection.
\newblock \emph{arXiv preprint arXiv:1611.02644}, 2016{\natexlab{b}}.

\bibitem[Liu et~al.(2022)Liu, Fan, Huang, Wu, Liu, Zhong, and Luo]{liu2022target}
Jinyuan Liu, Xin Fan, Zhanbo Huang, Guanyao Wu, Risheng Liu, Wei Zhong, and Zhongxuan Luo.
\newblock Target-aware dual adversarial learning and a multi-scenario multi-modality benchmark to fuse infrared and visible for object detection.
\newblock In \emph{Proceedings of the IEEE/CVF conference on computer vision and pattern recognition}, pages 5802--5811, 2022.

\bibitem[Liu et~al.(2021)Liu, Lam, Zhao, and Qiu]{liu2021deep}
Tianshan Liu, Kin-Man Lam, Rui Zhao, and Guoping Qiu.
\newblock Deep cross-modal representation learning and distillation for illumination-invariant pedestrian detection.
\newblock \emph{IEEE Transactions on Circuits and Systems for Video Technology}, 32\penalty0 (1):\penalty0 315--329, 2021.

\bibitem[Person et~al.(2019)Person, Jensen, Smith, and Gutierrez]{person2019multimodal}
Michael Person, Mathew Jensen, Anthony~O Smith, and Hector Gutierrez.
\newblock Multimodal fusion object detection system for autonomous vehicles.
\newblock \emph{Journal of Dynamic Systems, Measurement, and Control}, 141\penalty0 (7):\penalty0 071017, 2019.

\bibitem[Qingyun and Zhaokui(2022)]{qingyun2022cross}
Fang Qingyun and Wang Zhaokui.
\newblock Cross-modality attentive feature fusion for object detection in multispectral remote sensing imagery.
\newblock \emph{Pattern Recognition}, 130:\penalty0 108786, 2022.

\bibitem[Qingyun et~al.(2021)Qingyun, Dapeng, and Zhaokui]{qingyun2021cross}
Fang Qingyun, Han Dapeng, and Wang Zhaokui.
\newblock Cross-modality fusion transformer for multispectral object detection.
\newblock \emph{arXiv preprint arXiv:2111.00273}, 2021.

\bibitem[Ren et~al.(2017)Ren, He, Girshick, and Sun]{ren2017faster}
Shaoqing Ren, Kaiming He, Ross Girshick, and Jian Sun.
\newblock Faster r-cnn: Towards real-time object detection with region proposal networks.
\newblock \emph{IEEE Transactions on Pattern Analysis \& Machine Intelligence}, 39\penalty0 (06):\penalty0 1137--1149, 2017.

\bibitem[Shen et~al.(2024)Shen, Chen, Liu, Zuo, Fan, and Yang]{shen2024icafusion}
Jifeng Shen, Yifei Chen, Yue Liu, Xin Zuo, Heng Fan, and Wankou Yang.
\newblock Icafusion: Iterative cross-attention guided feature fusion for multispectral object detection.
\newblock \emph{Pattern Recognition}, 145:\penalty0 109913, 2024.

\bibitem[Shu et~al.(2021)Shu, Liu, Gao, Yan, and Shen]{shu2021channel}
Changyong Shu, Yifan Liu, Jianfei Gao, Zheng Yan, and Chunhua Shen.
\newblock Channel-wise knowledge distillation for dense prediction.
\newblock In \emph{Proceedings of the IEEE/CVF international conference on computer vision}, pages 5311--5320, 2021.

\bibitem[Sun et~al.(2022)Sun, Cao, Zhu, and Hu]{sun2022drone}
Yiming Sun, Bing Cao, Pengfei Zhu, and Qinghua Hu.
\newblock Drone-based rgb-infrared cross-modality vehicle detection via uncertainty-aware learning.
\newblock \emph{IEEE Transactions on Circuits and Systems for Video Technology}, 32\penalty0 (10):\penalty0 6700--6713, 2022.

\bibitem[Wang et~al.(2024)Wang, Chen, Zheng, Li, Cheng, and Hou]{wang2024crosskd}
Jiabao Wang, Yuming Chen, Zhaohui Zheng, Xiang Li, Ming-Ming Cheng, and Qibin Hou.
\newblock Crosskd: Cross-head knowledge distillation for object detection.
\newblock In \emph{Proceedings of the IEEE/CVF Conference on Computer Vision and Pattern Recognition}, pages 16520--16530, 2024.

\bibitem[Wolpert et~al.(2020)Wolpert, Teutsch, Sarfraz, and Stiefelhagen]{wolpert2020anchor}
Alexander Wolpert, Michael Teutsch, M~Saquib Sarfraz, and Rainer Stiefelhagen.
\newblock Anchor-free small-scale multispectral pedestrian detection.
\newblock \emph{arXiv preprint arXiv:2008.08418}, 2020.

\bibitem[Xia et~al.(2024)Xia, Wang, Lv, Hao, and Shi]{xia2024vit}
Chunlong Xia, Xinliang Wang, Feng Lv, Xin Hao, and Yifeng Shi.
\newblock Vit-comer: Vision transformer with convolutional multi-scale feature interaction for dense predictions.
\newblock In \emph{Proceedings of the IEEE/CVF conference on computer vision and pattern recognition}, pages 5493--5502, 2024.

\bibitem[Xu et~al.(2025)Xu, Xiang, Zhang, Zhong, Zhao, Dang, Xu, Pu, and Liu]{xu2024sckd}
Ruoyu Xu, Zhiyu Xiang, Chenwei Zhang, Hanzhi Zhong, Xijun Zhao, Ruina Dang, Peng Xu, Tianyu Pu, and Eryun Liu.
\newblock Sckd: Semi-supervised cross-modality knowledge distillation for 4d radar object detection.
\newblock \emph{Proceedings of the AAAI Conference on Artificial Intelligence}, 2025.

\bibitem[Yang et~al.(2022)Yang, Ochal, Storkey, and Crowley]{yang2022prediction}
Chenhongyi Yang, Mateusz Ochal, Amos Storkey, and Elliot~J Crowley.
\newblock Prediction-guided distillation for dense object detection.
\newblock In \emph{European Conference on Computer Vision}, pages 123--138. Springer, 2022.

\bibitem[Yang et~al.(2021)Yang, Zhang, Li, and Xie]{yang2021simam}
Lingxiao Yang, Ru-Yuan Zhang, Lida Li, and Xiaohua Xie.
\newblock Simam: A simple, parameter-free attention module for convolutional neural networks.
\newblock In \emph{International conference on machine learning}, pages 11863--11874. PMLR, 2021.

\bibitem[Yuan and Wei(2024)]{yuan2024c}
Maoxun Yuan and Xingxing Wei.
\newblock C 2 former: Calibrated and complementary transformer for rgb-infrared object detection.
\newblock \emph{IEEE Transactions on Geoscience and Remote Sensing}, 2024.

\bibitem[Yuan et~al.(2022)Yuan, Wang, and Wei]{yuan2022translation}
Maoxun Yuan, Yinyan Wang, and Xingxing Wei.
\newblock Translation, scale and rotation: cross-modal alignment meets rgb-infrared vehicle detection.
\newblock In \emph{European Conference on Computer Vision}, pages 509--525. Springer, 2022.

\bibitem[Yuan et~al.(2024{\natexlab{a}})Yuan, Cui, Zhao, and Wei]{yuan2024unirgb}
Maoxun Yuan, Bo Cui, Tianyi Zhao, and Xingxing Wei.
\newblock Unirgb-ir: A unified framework for visible-infrared downstream tasks via adapter tuning.
\newblock \emph{arXiv preprint arXiv:2404.17360}, 2024{\natexlab{a}}.

\bibitem[Yuan et~al.(2024{\natexlab{b}})Yuan, Shi, Wang, Wang, and Wei]{yuan2024improving}
Maoxun Yuan, Xiaorong Shi, Nan Wang, Yinyan Wang, and Xingxing Wei.
\newblock Improving rgb-infrared object detection with cascade alignment-guided transformer.
\newblock \emph{Information Fusion}, page 102246, 2024{\natexlab{b}}.

\bibitem[Zhang et~al.(2020)Zhang, Fromont, Lefevre, and Avignon]{9191080}
Heng Zhang, Elisa Fromont, Sébastien Lefevre, and Bruno Avignon.
\newblock Multispectral fusion for object detection with cyclic fuse-and-refine blocks.
\newblock In \emph{2020 IEEE International Conference on Image Processing (ICIP)}, pages 276--280, 2020.

\bibitem[Zhang et~al.(2021{\natexlab{a}})Zhang, Fromont, Lef{\`e}vre, and Avignon]{zhang2021guided}
Heng Zhang, Elisa Fromont, S{\'e}bastien Lef{\`e}vre, and Bruno Avignon.
\newblock Guided attentive feature fusion for multispectral pedestrian detection.
\newblock In \emph{Proceedings of the IEEE/CVF winter conference on applications of computer vision}, pages 72--80, 2021{\natexlab{a}}.

\bibitem[Zhang et~al.(2022)Zhang, Gu, Wang, Pan, and Wang]{zhang2022lane}
Han Zhang, Yunchao Gu, Xinliang Wang, Junjun Pan, and Minghui Wang.
\newblock Lane detection transformer based on multi-frame horizontal and vertical attention and visual transformer module.
\newblock In \emph{European Conference on Computer Vision}, pages 1--16. Springer, 2022.

\bibitem[Zhang et~al.(2023{\natexlab{a}})Zhang, Liu, Yang, Hu, Liu, and Stiefelhagen]{zhang2023cmx}
Jiaming Zhang, Huayao Liu, Kailun Yang, Xinxin Hu, Ruiping Liu, and Rainer Stiefelhagen.
\newblock Cmx: Cross-modal fusion for rgb-x semantic segmentation with transformers.
\newblock \emph{IEEE Transactions on intelligent transportation systems}, 24\penalty0 (12):\penalty0 14679--14694, 2023{\natexlab{a}}.

\bibitem[Zhang et~al.(2024)Zhang, Cao, Xie, Lei, Huang, Li, Yang, et~al.]{zhange2e}
Jiaqing Zhang, Mingxiang Cao, Weiying Xie, Jie Lei, Wenbo Huang, Yunsong Li, Xue Yang, et~al.
\newblock E2e-mfd: Towards end-to-end synchronous multimodal fusion detection.
\newblock In \emph{The Thirty-eighth Annual Conference on Neural Information Processing Systems}, 2024.

\bibitem[Zhang et~al.(2019)Zhang, Zhu, Chen, Yang, Lei, and Liu]{zhang2019weakly}
Lu Zhang, Xiangyu Zhu, Xiangyu Chen, Xu Yang, Zhen Lei, and Zhiyong Liu.
\newblock Weakly aligned cross-modal learning for multispectral pedestrian detection.
\newblock In \emph{Proceedings of the IEEE/CVF international conference on computer vision}, pages 5127--5137, 2019.

\bibitem[Zhang et~al.(2021{\natexlab{b}})Zhang, Liu, Zhu, Song, Yang, Lei, and Qiao]{zhang2021weakly}
Lu Zhang, Zhiyong Liu, Xiangyu Zhu, Zhan Song, Xu Yang, Zhen Lei, and Hong Qiao.
\newblock Weakly aligned feature fusion for multimodal object detection.
\newblock \emph{IEEE Transactions on Neural Networks and Learning Systems}, 2021{\natexlab{b}}.

\bibitem[Zhang et~al.(2023{\natexlab{b}})Zhang, Wang, Wang, Pang, Lyu, Zhang, Luo, and Chen]{zhang2023dense}
Shilong Zhang, Xinjiang Wang, Jiaqi Wang, Jiangmiao Pang, Chengqi Lyu, Wenwei Zhang, Ping Luo, and Kai Chen.
\newblock Dense distinct query for end-to-end object detection.
\newblock In \emph{Proceedings of the IEEE/CVF Conference on Computer Vision and Pattern Recognition}, pages 7329--7338, 2023{\natexlab{b}}.

\bibitem[Zhao et~al.(2024)Zhao, Yuan, Jiang, Wang, and Wei]{zhao2024removal}
Tianyi Zhao, Maoxun Yuan, Feng Jiang, Nan Wang, and Xingxing Wei.
\newblock Removal and selection: Improving rgb-infrared object detection via coarse-to-fine fusion.
\newblock \emph{arXiv preprint arXiv:2401.10731}, 2024.

\bibitem[Zhixing et~al.(2021)Zhixing, Zhang, Chang, Liu, Chen, Chen, et~al.]{zhixing2021distilling}
Du Zhixing, Rui Zhang, Ming Chang, Shaoli Liu, Tianshi Chen, Yunji Chen, et~al.
\newblock Distilling object detectors with feature richness.
\newblock \emph{Advances in Neural Information Processing Systems}, 34:\penalty0 5213--5224, 2021.

\end{thebibliography}
}

\end{document}